\documentclass[runningheads]{llncs}
\usepackage[T1]{fontenc}
\usepackage{graphicx}
\usepackage{booktabs}
\usepackage{array}
\usepackage{subfigure}
\usepackage{caption}
\usepackage{float}
\usepackage[misc]{ifsym}
\usepackage[colorlinks]{hyperref}
\usepackage{multirow}
\usepackage{url}
\urldef{\mailsa}\path|chunfeng.lian@xjtu.edu.cn|
\urldef{\mailsb}\path|fan.wang@xjtu.edu.cn|
\usepackage{booktabs}

\begin{document}
\title{\emph{Neuro}Explainer: Fine-Grained Attention Decoding to Uncover Cortical Development Patterns of Preterm Infants}
\titlerunning{\emph{Neuro}Explainer}
%


\author{Chenyu Xue\inst{1} \and Fan Wang\inst{2} \and Yuanzhuo Zhu\inst{2} \and Hui Li\inst{3} \and Deyu Meng\inst{1} \and Dinggang Shen\inst{4} \and Chunfeng Lian\inst{1}}



\institute{School of Mathematics and Statistics,Xi'an Jiaotong University, Xi'an, China\\
\mailsa \and
Key Laboratory of Biomedical Information Engineering of Ministry of Education,  School of Life Science and Technology, Xi'an Jiaotong University, Xi'an, China\\
\mailsb
 \and Department of Neonatology, The First Affiliated Hospital of Xi'an Jiaotong University, Xi'an, China\and School of Biomedical Engineering, ShanghaiTech University, Shanghai, China}

\maketitle              
\begin{abstract}
Deploying reliable deep learning techniques in interdisciplinary applications needs learned models to output accurate and (\emph{even more importantly}) explainable predictions. Existing approaches typically explicate network outputs in a post-hoc fashion,  under an implicit assumption that faithful explanations come from accurate predictions/classifications.
We have an opposite claim that explanations boost (or even determine) classification. That is, end-to-end learning of explanation factors to augment discriminative representation extraction could be a more intuitive strategy to inversely assure fine-grained explainability, e.g., in those neuroimaging and neuroscience studies with high-dimensional data containing noisy, redundant, and task-irrelevant information.
In this paper, we propose such an \emph{explainable geometric deep network} dubbed as \emph{Neuro}Explainer, with applications to uncover altered infant cortical development patterns associated with preterm birth. Given fundamental cortical attributes as network input, our \emph{Neuro}Explainer adopts a hierarchical attention-decoding framework to learn fine-grained attentions and respective discriminative representations to accurately recognize preterm infants from term-born infants at term-equivalent age. \emph{Neuro}Explainer learns the hierarchical attention-decoding modules under subject-level weak supervision coupled with \emph{targeted regularizers} deduced from domain knowledge regarding brain development. These prior-guided constraints implicitly maximizes the explainability metrics (i.e., fidelity, sparsity, and stability) in network training, driving the learned network to output detailed explanations and accurate classifications.
Experimental results on the public dHCP benchmark suggest that \emph{Neuro}Explainer led to quantitatively reliable explanation results that are qualitatively consistent with representative neuroimaging studies.

\end{abstract}
\section{Introduction}\label{sec:intro}
Due to the capacity of learning highly nonlinear representations in task-oriented fashions, deep neural networks are showing promising applications in many interdisciplinary communities, including biomedical image computing, neuroimaging, and neuroscience studies \cite{shen2017deep}.
To quantitatively analyze brain development/degeneration or timely diagnose associated disorders, various deep learning methods have been proposed \cite{yang2021deep,ouyang2022self}.
Most of these studies focused on the designs of network architectures and learning strategies to produce accurate predictions/classifications.
A critical challenge is that the learned models typically lack computational interpretability and results' explainability.
From the aspect of practical usage, neuroimaging and neuroscience studies desire AI tools that can make accurate and (\emph{even more importantly}) \emph{explainable} predictions.
For example, a key value of a brain disease diagnosis network is to identify from high-dimensional neuroimage data individualized subtle changes leading to accurate classification, which can be clues for human experts to analyze disease heterogeneity \cite{yang2021deep}.
Also, the classification task to differentiate between preterm and term-born (or male and female) infants may not be practically meaningful; \emph{in contrast}, fine-grained differences on brain cortical surfaces, identified by the learned classification network, could be valuable factors for better understanding featured cortical development patterns of infants from different groups.

Recently, explainable and interpretable deep learning is being actively studied in the machine learning community, with obviously more works on gridded data (e.g., images) \cite{du2019techniques} than on non-Euclidean data (e.g., 3D meshes) \cite{yuan2022explainability}.
Accurate classification and faithful explanation are highly correlated and inseparable.
Existing methods typically adopt post-hoc techniques to explain a deep network \cite{yuan2022explainability}, which is first trained for a specific classification task, and then the underlying (sparse) correlations between its input and output are analyzed offline, e.g., by backpropagating prediction gradients to the shallow layers \cite{smilkov2017smoothgrad}.
Notably, such post-hoc approaches are established upon a common assumption that reliable explanations are the results caused by accurate predictions.
This assumption could work in general applications that have large-scale training data, while cannot always hold for neuroimaging and neuroscience research, where available data are typically small-sized and much more complex (e.g., high-resolution cortical surfaces containing noisy, highly redundant, and task-irrelevant information).

Distinct to existing post-hoc methods, we have an opposite claim that \emph{explainability boosts or even determines classification}, especially in those challenging tasks related to brain cortical development analyses based on high-dimensional neuroimaging data. The key is how to construct an end-to-end framework, where fine-grained explanation factors can be identified in a fully learnable fashion to enhance discriminative representation extraction and finally output accurate classification.
This paper presents such an \emph{explainable geometric deep network}, called \emph{Neuro}Explainer, with applications to uncover altered infant cortical development patterns associated with preterm birth.
\emph{Neuro}Explainer adopts high-resolution cortical attributes as the input to develop a hierarchical attention-decoding architecture.
In the framework of weakly supervised discriminative localization, our \emph{Neuro}Explainer is trained by minimizing general classification losses coupled with a set of constraints designed according to prior knowledge regarding brain development. These targeted regularizers drive the network to implicitly optimize the explainability metrics from multiple aspects (i.e., fidelity, sparsity, and stability), thus capturing fine-grained explanation factors to explicitly improve classification accuracies.
Experimental results on the public dHCP benchmark suggest that our \emph{Neuro}Explainer led to quantitatively reliable explanation results that are qualitatively consistent with representative neuroimaging studies, implying that it could be a practically useful AI tool for other related cortical surface-based neuroimaging studies.

\section{Related Work}

\subsubsection{Deep Learning for Cortical Surface Analyses.}
The human cerebral cortex is a highly folded and thin sheet of gray matter \cite{fischl2000measuring}. By leveraging structural magnetic resonance imaging (MRI), such complex topology can be rendered as a 3D mesh, with each vertex/cell presenting fundamental cortical attributes, e.g., cortical thickness, mean curvature, and average convexity.
For brain development/degeneration analyses, advanced geometric deep learning methods can be potentially applied to learning from cortical surface data a powerful and efficient classification/prediction model, either over the original 3D mesh or after mapping it onto a spherical surface to ease the computation \cite{zhao2019spherical}.
Although these existing studies suggested promising accuracies of geometric deep learning in multiple tasks (e.g., parcellation \cite{zhao2019spherical}, registration \cite{suliman2022deep}, and longitudinal prediction \cite{liu2019deep}), the learned models typically lack explainability and interpretability.

\vspace{-12pt}
\subsubsection{Explainability in Deep Neural Networks.}
The instance-level explanation approaches aim to study why a deep model makes a specific prediction for a given input, e.g., which part of the input contributes more to its output classification score.
Such input-dependent explanation methods can be roughly categorized into four branches, including gradient/feature-based, perturbation-based, decomposition-based, and surrogate-based methods~\cite{yuan2022explainability}.
As the most straightforward strategy, gradient/feature-based approaches have been actively studied in deep learning over gridded data~\cite{du2019techniques}, with some extensions to non-Euclidean cases (e.g., graphs)~\cite{yuan2022explainability,mahmood2022through}.
Currently,  research regarding explainable geometric deep learning on 3D meshes (like brain cortical surfaces) is still limited~\cite{ribeiro2022explainability}.
More importantly, existing explanation methods (in both gridded and non-Euclidean spaces) typically investigate network outputs in a post-hoc fashion~\cite{du2019techniques,yuan2022explainability,liu2021going},  assuming that faithful explanations are the results of accurate predictions/classifications.
In this paper, we have an opposite claim that leveraging domain knowledge to perform end-to-end learning of feature-based explanation factors for discriminative representation extraction (and classification performance enhancement) could be a more intuitive strategy to produce fine-grained explainability in challenging tasks.

\begin{figure}[t]
    \includegraphics[width=\textwidth]{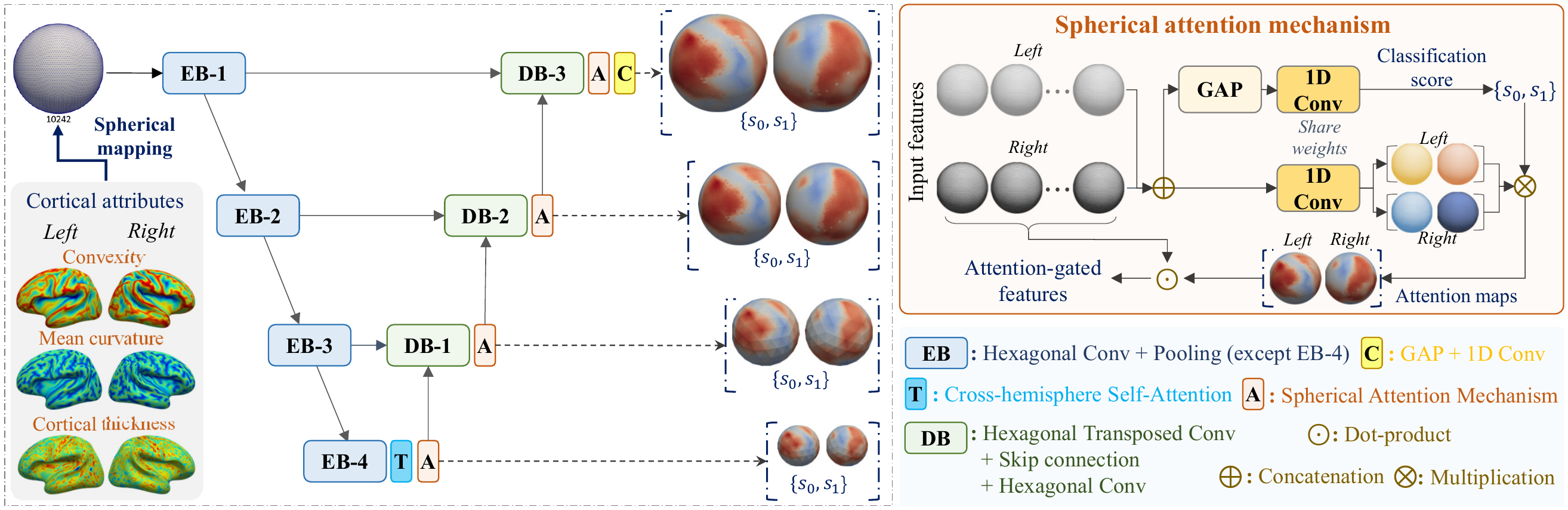}
    \caption{The schematic diagram of our \emph{Neuro}Explainer that learns to capture fine-grained explanation factors in an end-to-end attention-decoding architecture to boost discriminative representation extraction from cortical-surface data. } \label{fig:architecture}
\end{figure}

\section{Method} \label{sec:method}
As the schematic diagram shown in Fig.~\ref{fig:architecture}, our \emph{Neuro}Explainer works on the high-resolution spherical surfaces of both brain hemispheres (each with $10,242$ vertices).
The inputs are fundamental vertex-wise cortical attributes, i.e., thickness, mean curvature, and convexity.
The architecture has two main parts, including an encoding branch to produce initial task-related attentions on down-sampled hemispheric surfaces, and a set of attention decoding blocks to hierarchically propagate such vertex-wise attentions onto higher-resolution spheres, finally capturing fine-grained explanation factors on the input high-resolution surfaces to boost the prediction task (i.e., the differentiation between preterm and fullterm infants in our study).
The whole network is trained end-to-end by minimizing multi-resolution classification losses, under the constraints provided by prior-induced regularizations to enhance explanation metrics.

\subsection{Spherical Attention Encoding}\label{subsec:encoder}
The starting components of the encoding branch are four spherical convolution blocks (i.e., EB-1 to EB-4 in Fig.~\ref{fig:architecture}), with the learnable parameters shared across two hemispheric surfaces.
Each EB adopts 1-ring hexagonal convolution \cite{zhao2019spherical} followed by batch normalization (BN) and ReLU activation to extract vertex-wise representations, which are then downsampled by hexagonal max pooling \cite{zhao2019spherical} (except in EB-4) to serve as the input of the subsequent layer.
Based on the outputs from EB, we propose a learnable \emph{spherical attention mechanism} to conduct weakly-supervised discriminative localization.

Specifically, let $\mathbf{F}^l$ and $\mathbf{F}^r \in \mathcal{R}^{162\times M_0}$ be the vertex-wise representations (produced by EB-4) for the left and right hemispheres, respectively.
We first concatenate them as a $324\times  M_0$ matrix, on which a self-attention operation \cite{vaswani2017attention} is applied to capturing cross-hemisphere long-range dependencies to refine the vertex-wise representations from both hemispheric surfaces, resulting in a unified feature matrix denoted as $\mathbf{F_0}=[\hat{\mathbf{F}}^l; \hat{\mathbf{F}}^r] \in \mathcal{R}^{324\times M_0}$.
As shown in Fig. \ref{fig:architecture}, $\mathbf{F_0}$ is further global average pooled (GAP) across all vertices to be a holistic feature vector $f_0\in\mathcal{R}^{1\times M_0}$ representing the whole cerebral cortex. Both $\mathbf{F_0}$ and $f_0$ are then \emph{mapped by a same vertex-wise 1D convolution} (i.e., $\mathbf{W}_0\in \mathcal{R}^{M_0\times2}$, without bias) into the categorical space, denoted as $\mathbf{A}_0= [\mathbf{A}_0^l; \mathbf{A}_0^r] \in \mathcal{R}^{324\times 2}$ and $\mathbf{s}_0$, respectively.
\emph{Notably}, $\mathbf{s}_o$ is supervised by the one-hot code of subject's categorical label, by which $\mathbf{A}_0^l$ and $\mathbf{A}_0^r$ highlight discriminative vertices on the (down-sampled) left and right surfaces, respectively, considering that
\begin{equation}
\mathbf{s}_0[i] \propto\left({\mathbf{1}}^T\mathbf{F_0}\right)\mathbf{W}_0[:, i]= {\mathbf{1}}^T\left([\hat{\mathbf{F}}^l; \hat{\mathbf{F}}^r]\mathbf{W}_0[:, i]\right) ={\mathbf{1}}^T \mathbf{A}_0^l[:,i] + {\mathbf{1}}^T \mathbf{A}_0^r[:,i],
\label{equ:attention}
\end{equation}
where $\mathbf{s}_o[i]$ ($i=0$ or $1$) in our study denote the prediction scores of preterm and fullterm, respectively, and $\mathbf{1}$ is an unit vector having the same row size with the subsequent matrix. Finally, we define the hemispheric attentions as $\bar{\mathbf{A}}_0^l=\sum_{i=0}^1 \mathbf{s}_0[i]\mathbf{A}_0^l[:,i]$ and $\bar{\mathbf{A}}_0^r=\sum_{i=0}^1 \mathbf{s}_0[i]\mathbf{A}_0^r[:,i] \in \mathcal{R}^{324\times 1}$, respectively, with values spatially varying and depending on the relevance to subject's category.

\subsection{Hierarchically Spherical Attention Decoding} \label{subsec:decoder}
The explanation factors captured by the encoding branch are relatively coarse, as the receptive field of a cell on the downsampled surfaces (with 162 vertices after three pooling operations) is no smaller than a hexagonal region of 343 cells on the input surfaces (with $10,242$ vertices).
To tackle this challenge, we design a spherical attention decoding strategy to hierarchically propagate coarse attentions (from lower-resolution spheres) onto higher-resolution spheres, based on which fine-grained attentions are finally produced to improve classification.

Specifically, \emph{Neuro}Explainer contains three consecutive decoding blocks (i.e., DB-1 to DB-3 in Fig.~\ref{fig:architecture}).
Each DB adopts both the \emph{attention-gated} discriminative representations from the preceding DB (except DB-1 that uses EB-4 outputs) and the local-detailed representations from the symmetric EB (at the same resolution) as the input.
Let the attention-gated representations from the preceding DB be $\mathbf{F}_G^l=\left(\bar{\mathbf{A}}_{in}^l\mathbf{1}_{1\times M_{in}}\right)\odot \hat{\mathbf{F}}_{in}^l$ and $\mathbf{F}_G^r=\left(\bar{\mathbf{A}}_{in}^r\mathbf{1}_{1\times M_{in}}\right)\odot \hat{\mathbf{F}}_{in}^r$, respectively, where each row of $\hat{\mathbf{F}}_{in}$ has $M_{in}$ channels, and $\odot$ denotes element-wise dot product.
We first upsample $\mathbf{F}_G^l$ and $\mathbf{F}_G^r$ to the spatial resolution of the current DB, by using hexagonal transposed convolutions~\cite{zhao2019spherical} with learnable weights shared across hemispheres. Then, the upsampled discriminative representations from each hemisphere (say $\tilde{\mathbf{F}}_G^l$ and $\tilde{\mathbf{F}}_G^r$) are channel-wisely concatenated with the local representations from the corresponding EB (say $\mathbf{F}_E^l$ and $\mathbf{F}_E^r$), followed by an 1-ring convolution to produce a unified feature matrix, such as
\begin{equation}
\mathbf{F}_D = [\mathcal{C}_{\theta}(\tilde{\mathbf{F}}_G^l\oplus\mathbf{F}_E^l); \mathcal{C}_{\theta}(\tilde{\mathbf{F}}_G^r\oplus\mathbf{F}_E^r)],
\end{equation}
where $\mathcal{C}_{\theta}(\cdot)$ denotes 1-ring conv parameterized by $\theta$, and $\oplus$ stands for channel concatenation. In terms of $\mathbf{F}_D$, the attention mechanism described in~(\ref{equ:attention}) is further applied to producing refined spherical attentions and classification scores.

Finally, as shown in Fig.~\ref{fig:architecture}, based on the fine-grained attentions over the input surfaces (each with $10,242$ vertices), we use GAP to aggregate the attention-gated representations and apply an 1D conv to output the classification score.

\subsection{Domain Knowledge-Guided Explanation Enhancement} \label{subsec:regularization}
In this study, we design \emph{Neuro}Explainer under a central idea that task-oriented learning of explanation factors to boost discriminative representation extraction can inversely assure fine-grained explainability in challenging cases like learning on complex cortical-surface data.
To effectively train our network for such a purpose, we design a set of targeted regularization strategies by considering fundamental domain knowledge regarding infant brain development.
Specifically, it is reasonable to assume that human brains in infancy have generally consistent developments, while the structural/functional discrepancies between different groups (e.g., preterm and term-born) are typically localized~\cite{thompson2020tracking,dimitrova2021preterm}.
Accordingly, we require the preterm-altered cortical development patterns captured by our \emph{Neuro}Explainer to be discriminative, spatially sparse, and robust, which suggests the design of the following constraints that concurrently optimize fidelity, sparsity, and stability metrics \cite{yuan2022explainability} in deploying an explainable deep network.

\begin{figure}[t]
    \includegraphics[width=1\textwidth]{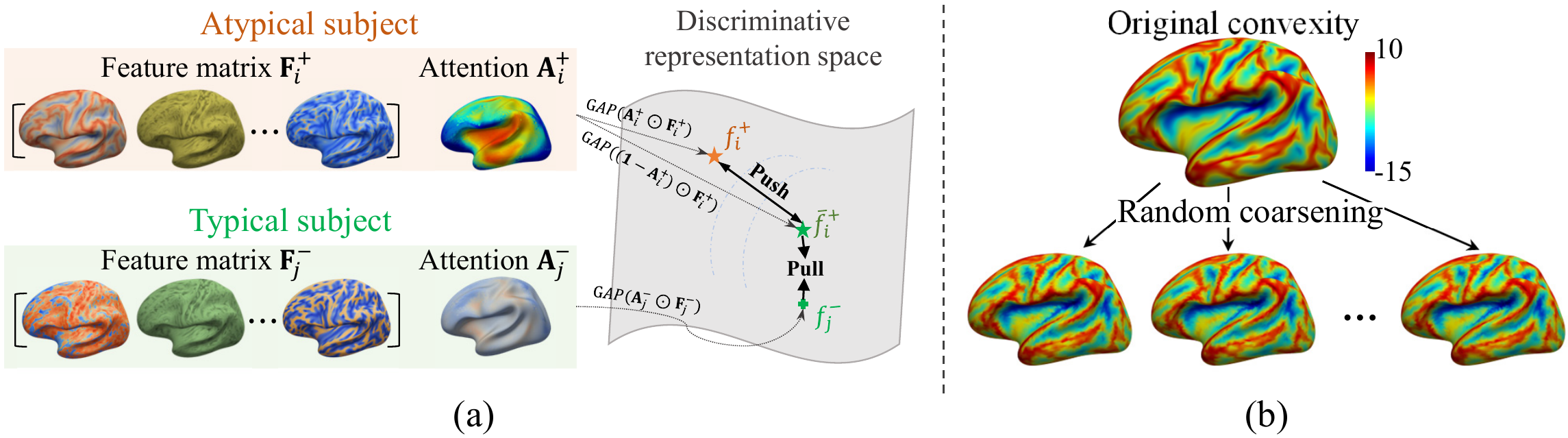}
    \caption{Brief illustrations of (a) the explanation fidelity-aware contrastive learning strategy, and (b) explanation stability-aware data augmentation strategy.} \label{fig:contrast}
\end{figure}

\vspace{-12pt}
\subsubsection{Explanation Fidelity-Aware Contrastive Learning.}
Given the spherical attention block at a specific resolution, we have $\mathbf{A}^+_i$ and $\mathbf{A}^-_j\in\mathcal{R}^{V\times1}$ as the output attentions for a positive and negative subjects (i.e., preterm and fullterm infants in our study), respectively, and $\mathbf{F}^+_i$ and $\mathbf{F}^-_j\in\mathcal{R}^{V\times M}$ are the corresponding representation matrices.
Based on the prior knowledge regarding infant brain development, it is reasonable to assume that
$\mathbf{A}^+_i$ highlights atypically-developed cortical regions caused by preterm birth.
\emph{In contrast}, the remaining part of the cerebral cortex of a preterm infant (corresponding to $1-\mathbf{A}^+_i$) still growths normally, i.e., looking globally similar to the cortex of a term-born infant.

Accordingly, as the illustration shown in Fig. \ref{fig:contrast}(a), we design a fidelity-aware contrastive penalty to regularize the learning of the attention maps and associated representations to improve their discriminative power. Let $f^+_i = \mathbf{1}^T\left(\mathbf{A}^+_i\mathbf{1}_{1\times M}\odot\mathbf{F}^+_i\right)$ and $\bar{f}^+_i = \mathbf{1}^T\left(\{1-\mathbf{A}^+_i\}\mathbf{1}_{1\times M}\odot\mathbf{F}^+_i\right)$ be the holistic feature vector and its inverse for the $i$th (positive) sample, respectively. Similarly, $f^-_j = \mathbf{1}^T\left(\mathbf{A}^-_j\mathbf{1}_{1\times M}\odot\mathbf{F}^-_j\right)$ denotes the holistic feature vector for the compared $j$th (negative) sample.
By \emph{pushing $f_i^+$ away from both $\bar{f}_i^+$ and $f_j^-$, while pulling $\bar{f}_i^+$ close to $f_j^-$}, we define the respective loss as
\begin{equation}
\mathcal{L}_{contrast} = \sum_{i\neq j}^N||\bar{f}_i^+-f_j^-|| + \max(m-||\bar{f}_i^+-f_i^+||,0)+\max(m-||f_j^--f_i^+||,0),
\label{equ:contrast}
\end{equation}
where $i$ and $j$ indicate any a pair of positive and negative cases from totally $N$ training samples, and $m$ is a margin setting as 1 in our implementation.

\vspace{-12pt}
\subsubsection{Explanation Sparsity-Aware Regularization.}
According to the specified prior knowledge regarding infant brain development, the attention maps produced by our \emph{Neuro}Explainer should have two featured properties in terms of sparsity.
That is, the attention map for a preterm infant (e.g., $\mathbf{A}^+_i$) should be sparse, considering that altered cortical developments are assumed to be localized.
In contrast, the attention map for a healthy term-born infant (e.g., $\mathbf{A}^-_j$) should not be spatially informative, as all brain regions growth typically without abnormality.
To this end, we design a straightforward entropy-based regularization to enhance results' explainability, such as
\begin{equation}
    \mathcal{L}_{entropy}=\sum_{i\neq j}^N \mathbf{1}^T\left\{\mathbf{A}_i^+\odot\log(\mathbf{A}_i^+)-\mathbf{A}_j^-\odot\log(\mathbf{A}_j^-)\right\},
\label{equ:entropy}
\end{equation}
where $i$ and $j$ indicate a positive and a negative cases from totally $N$ training samples, respectively, and $\mathbf{1}$ is an unit vector to sum up the values of all vertices.

\vspace{-12pt}
\subsubsection{Explanation Stability-Aware Regularization.}
We enhance the explanation stability of our \emph{Neuro}Explainer from two aspects. \emph{First}, we require the spherical attention mechanisms to \emph{robustly} decode from complex cortical-surface data (potentially containing perturbed and noisy information) fine-grained explanation factors to produce accurate predictions.
To this end, a customized data augmentation strategy is designed to explicitly increase perturbations and variances in preprocessing the data for training such a network.
Specifically, the inputs of our network are cortical surfaces with $10,242$ vertices that are downsampled from the original data with $163,842$ vertices.
We randomize this surface coarsening step by quantifying a vertex's cortical attributes (on the downsampled surface) as the average of a random subset of the vertices from the respective hexagonal region of the highest-resolution surface. As the examples summarized in Fig. \ref{fig:contrast}(b), such a data augmentation strategy can generate partially different training samples from one single subject.
Considering that the network is trained to produce consistently accurate predictions for all these variants with perturbations, it inversely enhances the stability of learned explanation factors.

\emph{Second}, as described in Sec. \ref{subsec:decoder}, the fine-grained explanation factors over high-resolution cortical surface are captured by the proposed hierarchical attention decoding strategy, where coarse results from the encoder part serve as the foundation. To further enhance the explanation stability, we design a cross-scale consistency regularization to refine the decoding branch.
Specifically, let $\mathbf{A}_i^{l}$ and $\mathbf{A}_i^{h}$ be the spherical attentions from two different DB blocks. We simply minimize
\begin{equation}
    \mathcal{L}_{consistent}=\sum_{i=1}^N{\left(\mathbf{A}_i^{l}-\mathbf{A}_i^{h} \right)}^2,
    \label{equ:consistent}
\end{equation}
which encourages spherical attention maps at different spatial resolutions to be consistent in network training.

\vspace{-12pt}
\subsubsection{Implementation Details.}
In our implementation, the feature representations produced by EB-1 to EB-4 in Fig. \ref{fig:architecture} have 32, 64, 128, and 256 channels, respectively. Correspondingly, DB-1 to DB-3, and the final classification layer have 256, 128,
64, and 32 channels, respectively.
The network was trained end-to-end by minimizing the cross-entropy classification losses defined at three different spatial resolutions (overall denoted as $\mathcal{L}_{CE}$), coupled with the regularization terms introduced in Sec. \ref{subsec:regularization}, such as
\begin{equation}
   \mathcal{L}=\mathcal{L}_{CE}+\lambda _1\mathcal{L}_{contrast}+\lambda _2\mathcal{L}_{entropy}+\lambda _3\mathcal{L}_{consistent},
   \label{equ:loss}
\end{equation}
where the tuning parameters were empirically set as $\lambda_1=0.2$, $\lambda_3=0.5$, and $\lambda_3=0.1$.
The network parameters were updated by using Adam optimizer for 500 epochs, with the initial learning rate setting as $0.001$ and bath size as $20$.

\section{Experiments}

\subsubsection{Dataset \& Experimental Setup.}
We conducted experiments on the dHCP benchmark~\cite{makropoulos2018developing}.
The structural MRIs of 700 infants scanned at term-equivalent ages (35-44 weeks postmenstrual age) were studied, including 143 preterm and 557 term-born infants.
These subjects were randomly split as a training set of 500 infants (89 preterm and 411 fullterm), and a test set of the remaining 200 infants (54 preterm and 146 fullterm).
Using the data-augmentation strategy described in Sec.~\ref{subsec:regularization}, the training set was augmented to have roughly $1,250$ subjects from each category for balanced network training.
The input spherical surfaces contain $10,242$ vertices, and each of them has three morphological attributes, i.e., cortical thickness, mean curvature, and convexity.

For classification, our \emph{Neuro}Explainer was compared with three representative geometric deep networks, including a spherical network based on 1-ring convolution (\textbf{SphericalCNN})~\cite{zhao2019spherical}, a MoNet reimplementation working on spherical surfaces (\textbf{SphericalMoNet})~\cite{suliman2022deep}, and \textbf{SubdivNet}~\cite{hu2022subdivision} working on original cortical meshes.
In addition, we conducted detailed ablation studies to verify the efficacy of each prior-guided regularization introduced in Sec.~\ref{subsec:regularization}. The classification performance was quantified in terms of accuracy (\textbf{ACC}), area under the ROC curve (\textbf{AUC}), sensitivity (\textbf{SEN}), and specificity (\textbf{SPE}).

On the other hand, the explanation performance of our \emph{Neuro}Explainer was compared with two representative feature-based explanation approaches, i.e., \textbf{CAM}~\cite{zhou2016learning} and \textbf{Grad-CAM}~\cite{selvaraju2017grad}, which were coupled with the geometric networks described above for post-hoc analysis. The explanation performance was quantitatively evaluated in terms of three metrics~\cite{yuan2022explainability}, i.e., \textbf{Fidelity} that measures the classification differences between the network captured explanation factors and the remaining part, \textbf{Sparsity} of the captured explanation factors compared to the whole surface, and \textbf{Stability} that measures the average classification accuracy on different perturbations of a single subject.
Please refer to~\cite{yuan2022explainability} for more details regarding these explainability evaluation metrics.

\begin{table}[t]
\setlength{\tabcolsep}{8pt}
\centering
\caption{Classification results obtained by the competing geometric deep networks and different variants of our \emph{Neuro}Explainer.}
 \begin{tabular}{r||m{1cm}<{\centering}m{1cm}<{\centering}m{1cm}
        <{\centering}m{1cm}<{\centering}}
    \hline
 \textbf{Competing Mehtods}    &  \textbf{ACC} & \textbf{AUC} & \textbf{SEN} & \textbf{SPE} \\
  \hline
  \hline
   SphericalCNN~\cite{zhao2019spherical} &  0.93 & 0.92 & 0.76 & \textbf{0.98}\\
   SphericalMoNet~\cite{suliman2022deep} &  0.85 & 0.93 & 0.65 & 0.92\\
   SubdivNet~\cite{hu2022subdivision} &  0.79 & 0.67 & 0.74 & 0.80\\
   \hline
   \hline
   \emph{Neuro}Explainer (\textbf{ours}) &  \textbf{0.95} & \textbf{0.97} & \textbf{0.94} & 0.95\\
  w/o $\mathcal{L}_{contrast}$ (\ref{equ:contrast}) & 0.88 & 0.89 & 0.80 & 0.91\\
   w/o $\mathcal{L}_{entropy}$ (\ref{equ:entropy})  & 0.91 & 0.96 & 0.74 & 0.97\\
   w/o $\mathcal{L}_{consistent}$ (\ref{equ:consistent})  & 0.89 & 0.95 & 0.89 & 0.88\\
\hline
    \end{tabular}
\label{tab:classification}
\end{table}

\vspace{-12pt}
\subsubsection{Classification Results.}
The classification results obtained by different competing methods are summarized in Table~\ref{tab:classification}, from which we can have at least \emph{three observations}.
\textbf{1)} Compared with SubdivNet working on original meshes, and the other two networks working on spherical surfaces (i.e., SphericalCNN and SphericalMoNet), our \emph{Neuro}Explainer consistently led to better classification accuracies in terms of all metrics.
\textbf{2)} The improvements brought by our method are especially significant in terms of SEN and AUC, suggesting that it can reliably identify featured development patterns associated with preterm birth to make accurate predictions in such an imbalanced learning task.
These results imply that our idea to capture fine-grained explanation factors in an end-to-end fashion to boost discriminative representation extraction is beneficial for deploying an accurate classification model in the task of learning on complex surface data containing noisy, redundant, and task-irrelevant information.

\textbf{3)} To check the efficacy of the prior-induced regularization strategies, we orderly removed them from the loss function~(\ref{equ:loss}) to quantify the respective influence on classification results.
From Table~\ref{tab:classification}, we can see that all the three regularizations (i.e., $\mathcal{L}_{contrast}$, $\mathcal{L}_{entropy}$, and $\mathcal{L}_{consistent}$) demonstrated \emph{significant but different} improvements on classification. Specifically, according to the comparison between \emph{Neuro}Explainer and its variant w/o $\mathcal{L}_{entropy}$, we can see that such an explanation sparsity-aware regularization boosted SEN by 20\%, implying its efficacy in capturing localized patterns related to preterm-altered cortical development. By comparing \emph{Neuro}Explainer with its variant w/o $\mathcal{L}_{consistent}$, we can see that the explanation stability-ware regularization helped stabilize the learning of the hierarchical attention-decoding branch, leading to relatively large improvements of overall classification ACC (by 6\%). Finally, we can see that the explanation fidelity-aware contrastive learning strategy brought overall the largest improvements of both ACC and AUC (by 7\% and 8\%, respectively), implying its efficacy in transforming domain knowledge regarding atypical brain development to capture fine-grained explanation factors that boost discriminative representation learning.

\begin{table}[t]
\centering
\caption{Quantitative explanation results obtained by the competing post-hoc approaches and our end-to-end \emph{Neuro}Explainer.}
\begin{tabular}{rl||m{1.5cm}<{\centering}m{1.5cm}<{\centering}m{1.5cm}
        <{\centering}m{1.5cm}<{\centering}}
\hline
\multicolumn{2}{c||}{\textbf{Competing Methods}} & \textbf{Fidelity} & \textbf{Sparsity} & \textbf{Stability}\\
\hline
\hline
\multirow{3}{*}{CAM~\cite{zhou2016learning} + } &  SphericalCNN &  0.24 &  0.91 &  0.77 \\
                                            &  SphericalMoNet &  0.55 &  0.93 &  0.58 \\
                                            &       SubdivNet   &  0.06 &  0.97 &  0.53 \\
\hline
\multirow{3}{*}{Grad-CAM~\cite{selvaraju2017grad} + } &  SphericalCNN &  0.22 &  0.99 &  0.77 \\
                                            &  SphericalMoNet &  0.42 &  0.98 &  0.58 \\
                                            &       SubdivNet   &  0.16 &  0.96 &  0.53 \\
\hline
\hline
\multicolumn{2}{c||}{\emph{Neuro}Explainer (\textbf{ours})} & \textbf{0.56} &  0.73 &  \textbf{0.96} \\
\hline
\end{tabular}
\label{tab:explanation}
\end{table}

\begin{figure}[t]
    \centering
    \includegraphics[width=1\textwidth]{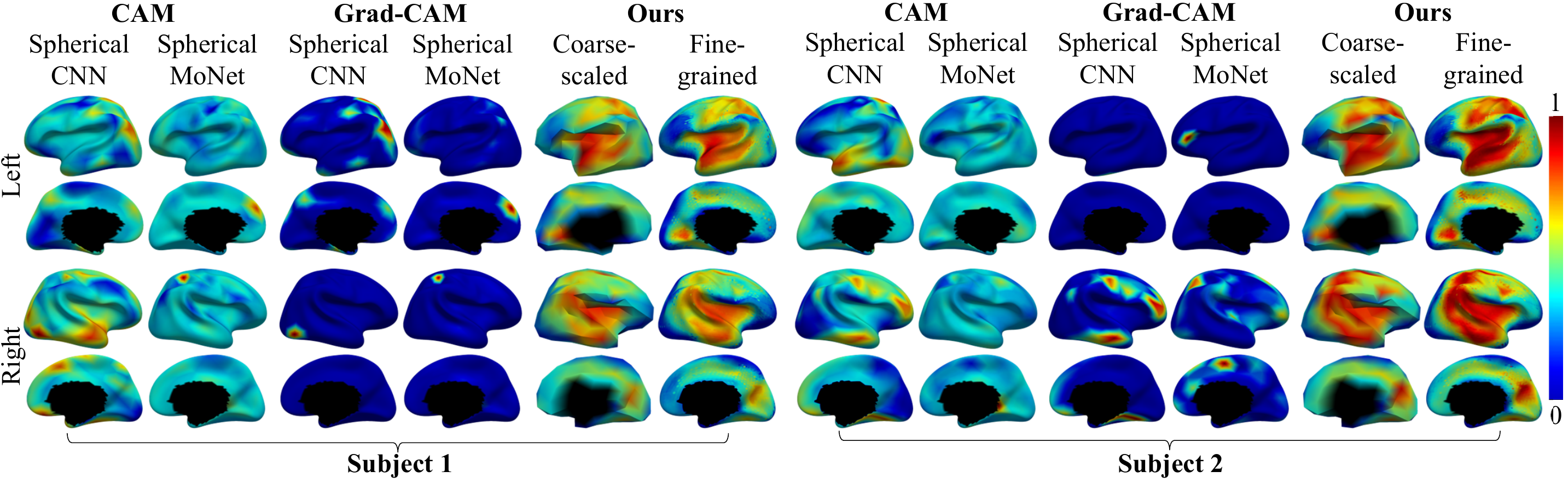}
    \caption{Typical examples of the explanation factors captured by the competing post-hoc approaches (CAM and Grad-CAM) and our end-to-end \emph{Neuro}Explainer. Higher values indicate larger links to preterm birth.}
     \label{fig:attention}
\end{figure}

\vspace{-12pt}
\subsubsection{Explanation Results.}
The quantitative explanation results obtained by our \emph{Neuro}Explainer and other feature-based explanation methods (i.e., CAM and Grad-CAM coupled with the trained geometric deep networks, respectively) are summarized in Table~\ref{tab:explanation}.
From Table~\ref{tab:explanation}, we can observe that our end-to-end \emph{Neuro}Explainer outperformed other post-hoc explanation approaches by a large margin in terms of the three explainability metrics.
Notably, the three metrics should be analyzed concurrently in evaluating a network's explainability~\cite{yuan2022explainability}, as the isolated quantification of a single metric could be biased.
For example, although Grad-CAM+SphericalCNN led to the largest Sparsity value ($=0.99$) in our experiment, the corresponding Fidelity and Stability values are significantly low ($=0.22$ and $0.77$, respectively), indicating that the very sparse explanation factors captured by Grad-CAM+SphericalCNN are relatively uninformative and random.
\emph{In contrast}, our \emph{Neuro}Explainer led to significantly better Fidelity and Stability, under reasonable Sparsity, suggesting that it can robustly identify localized preterm-altered cortical patterns from high-dimensional inputs to boost discriminative representation learning for preterm infant recognition.

In addition to the above quantitative evaluations, we also visually compared the attention maps produced by different competing methods, with two typical examples presented in Fig.~\ref{fig:attention}.
From Fig.~\ref{fig:attention}, we can have \emph{two main observations}. \textbf{1)} Compared with post-hoc explanation methods (i.e., CAM and Grad-CAM), our end-to-end \emph{Neuro}Explainer stably produced more reasonable attentions.
Specifically, given a subject (e.g., Subject 2 in Fig.~\ref{fig:attention}), CAM and Grad-CAM could produce very different explanation results for a trained classification model (e.g., SphericalCNN). Similarly, a post-hoc approach (e.g., CAM) could produce distinct results for two different classification models on the same subject.
Due to the nature of end-to-end learning of explanation factors to establish classification models, our \emph{Neuro}Explainer effectively avoided such a problem. \emph{More importantly}, across different subjects, our \emph{Neuro}Explainer led to group-wisely more consistent explanations than these post-hoc approaches. Also, it produced more consistent results across hemispheres, without any related constraints during network training.
\textbf{2)} We can see that the coarse attentions produced by \emph{Neuro}Explainer's encoder are consistent with the final fine-grained outputs of the decoder, which implies the positive effect of the cross-scale consistency regularization in enhancing explainability.

Finally, we compared the \emph{individualized} preterm-altered cortical development patterns uncovered by our \emph{Neuro}Explainer with representative \emph{group-wise} neuroimaging studies in the literature, e.g., the multi-modal (dMRI and sMRI) quantitative analyses presented in~\cite{dimitrova2021preterm}.
According to the comparisons shown in Fig.~\ref{fig:attention}, we can see that our observations in this paper are consistent with~\cite{dimitrova2021preterm}.
The discriminative cortical regions captured by our \emph{Neuro}Explainer (using solely morphological features) are largely overlapped with the group-wise significantly different regions identified by~\cite{dimitrova2021preterm} in terms of the mean diffusivity, neurite density, and cortical thickness, respectively.
For example, they both highlighted some specific regions in the inferior parietal, medial occipital, and superior temporal lobe, and posterior insula, which is worth deeper evaluations in the future.

\begin{figure}[t]
  \centering
    \includegraphics[width=1\textwidth]{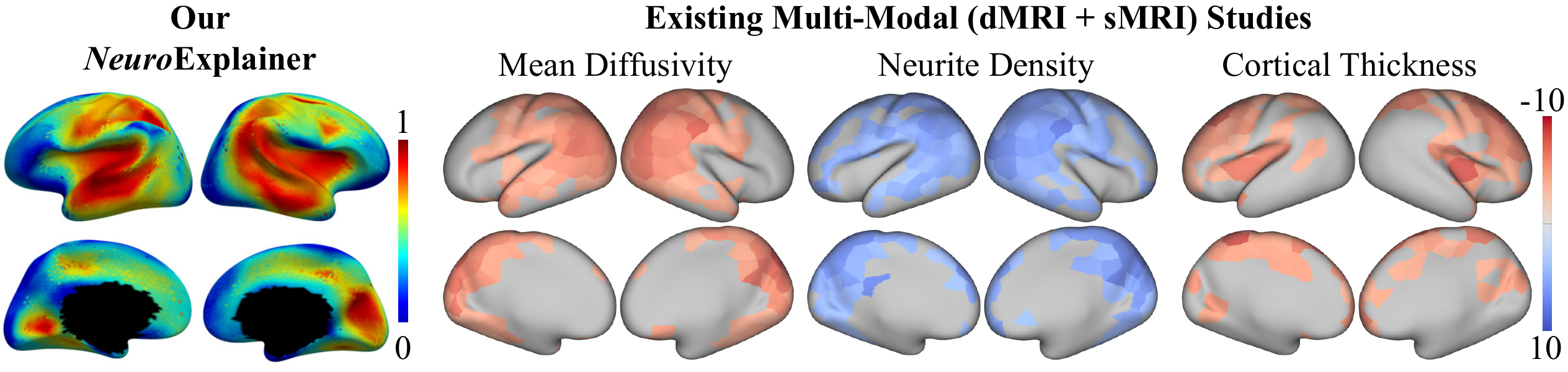}
    \caption{Comparison of the \emph{individualized} preterm-altered cortical development patterns uncovered by \emph{Neuro}Explainer with the \emph{group-wise} patterns identified by existing multi-modal studies~\cite{dimitrova2021preterm}.}
    \label{fig:neuroscience}
\end{figure}

\section{Conclusion}

In the paper, we have proposed an geometric deep network, i.e., \emph{Neuro}Explainer, to learn fine-grained explanation factors from complex cortical-surface data to boost discriminative representation extraction and accurate classification model construction. On the benchmark dHCP database, the applications of our \emph{Neuro}Explainer to uncover preterm-altered infant cortical development patterns achieved better performance in terms of both explainability and prediction accuracy, when compared with representative post-hoc approaches coupled with state-of-the-art geometric deep networks. The proposed method could be a promising AI tool applied to other similar cortical surface-based neuroimage and neuroscience studies.

%
%
%
\bibliographystyle{splncs04}
\bibliography{ref-ipmi}

\begin{thebibliography}{10}
\providecommand{\url}[1]{\texttt{#1}}
\providecommand{\urlprefix}{URL }
\providecommand{\doi}[1]{https://doi.org/#1}

\bibitem{ball2020cortical}
Ball, G., Seidlitz, J., O'Muircheartaigh, J., Dimitrova, R., Fenchel, D.,
  Makropoulos, A., Christiaens, D., Schuh, A., Passerat-Palmbach, J., Hutter,
  J., et~al.: Cortical morphology at birth reflects spatiotemporal patterns of
  gene expression in the fetal human brain. PLoS biology  \textbf{18}(11),
  e3000976 (2020)

\bibitem{ball2013development}
Ball, G., Srinivasan, L., Aljabar, P., Counsell, S.J., Durighel, G., Hajnal,
  J.V., Rutherford, M.A., Edwards, A.D.: Development of cortical microstructure
  in the preterm human brain. Proceedings of the National Academy of Sciences
  \textbf{110}(23),  9541--9546 (2013)

\bibitem{bass2022icam}
Bass, C., et~al.: Icam-reg: Interpretable classification and regression with
  feature attribution for mapping neurological phenotypes in individual scans.
  IEEE TMI  (2022)

\bibitem{cheplygina2019not}
Cheplygina, V., de~Bruijne, M., Pluim, J.P.: Not-so-supervised: a survey of
  semi-supervised, multi-instance, and transfer learning in medical image
  analysis. Medical image analysis  \textbf{54},  280--296 (2019)

\bibitem{cirecsan2013mitosis}
Cire{\c{s}}an, D.C., Giusti, A., Gambardella, L.M., Schmidhuber, J.: Mitosis
  detection in breast cancer histology images with deep neural networks. In:
  International conference on medical image computing and computer-assisted
  intervention. pp. 411--418. Springer (2013)

\bibitem{cui2022interpretable}
Cui, H., et~al.: Interpretable graph neural networks for connectome-based brain
  disorder analysis. In: MICCAI. pp. 375--385. Springer (2022)

\bibitem{davis1987behavioral}
Davis, D.H., Thoman, E.B.: Behavioral states of premature infants: Implications
  for neural and behavioral development. Developmental Psychobiology
  \textbf{20}(1),  25--38 (1987)

\bibitem{de2016machine}
De~Bruijne, M.: Machine learning approaches in medical image analysis: From
  detection to diagnosis (2016)

\bibitem{dietterich1997solving}
Dietterich, T.G., Lathrop, R.H., Lozano-P{\'e}rez, T.: Solving the multiple
  instance problem with axis-parallel rectangles. Artificial intelligence
  \textbf{89}(1-2),  31--71 (1997)

\bibitem{dimitrova2021preterm}
Dimitrova, R., et~al.: Preterm birth alters the development of cortical
  microstructure and morphology at term-equivalent age. NeuroImage
  \textbf{243},  118488 (2021)

\bibitem{du2019techniques}
Du, M., et~al.: Techniques for interpretable machine learning. Communications
  of the ACM  \textbf{63}(1),  68--77 (2019)

\bibitem{fawaz2021benchmarking}
Fawaz, A., et~al.: Benchmarking geometric deep learning for cortical
  segmentation and neurodevelopmental phenotype prediction. bioRxiv  (2021)

\bibitem{feng2019meshnet}
Feng, Y., et~al.: Meshnet: Mesh neural network for 3d shape representation. In:
  AAAI. vol.~33, pp. 8279--8286 (2019)

\bibitem{fischl2012freesurfer}
Fischl, B.: Freesurfer. Neuroimage  \textbf{62}(2),  774--781 (2012)

\bibitem{fischl2000measuring}
Fischl, B., et~al.: Measuring the thickness of the human cerebral cortex from
  magnetic resonance images. PNAS  \textbf{97}(20),  11050--11055 (2000)

\bibitem{hu2022subdivision}
Hu, S.M., et~al.: Subdivision-based mesh convolution networks. ACM TOG
  \textbf{41}(3),  1--16 (2022)

\bibitem{huang2019evidence}
Huang, Y., Chung, A.: Evidence localization for pathology images using weakly
  supervised learning. In: International conference on medical image computing
  and computer-assisted intervention. pp. 613--621. Springer (2019)

\bibitem{kostovic2006development}
Kostovi{\'c}, I., Jovanov-Milo{\v{s}}evi{\'c}, N.: The development of cerebral
  connections during the first 20--45 weeks' gestation. In: seminars in fetal
  and neonatal medicine. vol.~11, pp. 415--422. Elsevier (2006)

\bibitem{krizhevsky2017imagenet}
Krizhevsky, A., Sutskever, I., Hinton, G.E.: Imagenet classification with deep
  convolutional neural networks. Communications of the ACM  \textbf{60}(6),
  84--90 (2017)

\bibitem{li2012consistent}
Li, G., et~al.: Consistent reconstruction of cortical surfaces from
  longitudinal brain mr images. NeuroImage  \textbf{59}(4),  3805--3820 (2012)

\bibitem{litjens2017survey}
Litjens, G., Kooi, T., Bejnordi, B.E., Setio, A.A.A., Ciompi, F., Ghafoorian,
  M., Van Der~Laak, J.A., Van~Ginneken, B., S{\'a}nchez, C.I.: A survey on deep
  learning in medical image analysis. Medical image analysis  \textbf{42},
  60--88 (2017)

\bibitem{liu2019deep}
Liu, P., et~al.: Deep modeling of growth trajectories for longitudinal
  prediction of missing infant cortical surfaces. In: IPMI. pp. 277--288.
  Springer (2019)

\bibitem{liu2021going}
Liu, Z., et~al.: Going beyond saliency maps: Training deep models to interpret
  deep models. In: IPMI. pp. 71--82. Springer (2021)

\bibitem{mahmood2022through}
Mahmood, U., et~al.: Through the looking glass: deep interpretable dynamic
  directed connectivity in resting fmri. NeuroImage p. 119737 (2022)

\bibitem{makropoulos2016regional}
Makropoulos, A., Aljabar, P., Wright, R., H{\"u}ning, B., Merchant, N., Arichi,
  T., Tusor, N., Hajnal, J.V., Edwards, A.D., Counsell, S.J., et~al.: Regional
  growth and atlasing of the developing human brain. Neuroimage  \textbf{125},
  456--478 (2016)

\bibitem{makropoulos2018developing}
Makropoulos, A., et~al.: The developing human connectome project: A minimal
  processing pipeline for neonatal cortical surface reconstruction. NeuroImage
  \textbf{173},  88--112 (2018)

\bibitem{monti2017geometric}
Monti, F., Boscaini, D., Masci, J., Rodola, E., Svoboda, J., Bronstein, M.M.:
  Geometric deep learning on graphs and manifolds using mixture model cnns. In:
  Proceedings of the IEEE conference on computer vision and pattern
  recognition. pp. 5115--5124 (2017)

\bibitem{ouyang2022self}
Ouyang, J., et~al.: Self-supervised learning of neighborhood embedding for
  longitudinal mri. MedIA  \textbf{82},  102571 (2022)

\bibitem{quellec2017multiple}
Quellec, G., Cazuguel, G., Cochener, B., Lamard, M.: Multiple-instance learning
  for medical image and video analysis. IEEE reviews in biomedical engineering
  \textbf{10},  213--234 (2017)

\bibitem{ribeiro2022explainability}
Ribeiro, F.L., et~al.: An explainability framework for cortical surface-based
  deep learning. arXiv preprint arXiv:2203.08312  (2022)

\bibitem{selvaraju2017grad}
Selvaraju, R.R., et~al.: Grad-cam: Visual explanations from deep networks via
  gradient-based localization. In: CVPR. pp. 618--626 (2017)

\bibitem{seong2018geometric}
Seong, S.B., Pae, C., Park, H.J.: Geometric convolutional neural network for
  analyzing surface-based neuroimaging data. Frontiers in neuroinformatics
  \textbf{12}, ~42 (2018)

\bibitem{shen2017deep}
Shen, D., et~al.: Deep learning in medical image analysis. Annual Review of
  Biomedical Engineering  \textbf{19}, ~221 (2017)

\bibitem{smilkov2017smoothgrad}
Smilkov, D., et~al.: Smoothgrad: removing noise by adding noise. arXiv preprint
  arXiv:1706.03825  (2017)

\bibitem{srinidhi2021deep}
Srinidhi, C.L., Ciga, O., Martel, A.L.: Deep neural network models for
  computational histopathology: A survey. Medical Image Analysis  \textbf{67},
  101813 (2021)

\bibitem{suliman2022deep}
Suliman, M.A., et~al.: A deep-discrete learning framework for spherical surface
  registration. In: MICCAI. pp. 119--129. Springer (2022)

\bibitem{thompson2020tracking}
Thompson, D.K., et~al.: Tracking regional brain growth up to age 13 in children
  born term and very preterm. Nature Communications  \textbf{11}(1),  1--11
  (2020)

\bibitem{vaswani2017attention}
Vaswani, A., et~al.: Attention is all you need. NeurIPS  \textbf{30} (2017)

\bibitem{volpe2019dysmaturation}
Volpe, J.J.: Dysmaturation of premature brain: importance, cellular mechanisms,
  and potential interventions. Pediatric neurology  \textbf{95},  42--66 (2019)

\bibitem{wang2019rmdl}
Wang, S., Zhu, Y., Yu, L., Chen, H., Lin, H., Wan, X., Fan, X., Heng, P.A.:
  Rmdl: Recalibrated multi-instance deep learning for whole slide gastric image
  classification. Medical image analysis  \textbf{58},  101549 (2019)

\bibitem{weese2016four}
Weese, J., Lorenz, C.: Four challenges in medical image analysis from an
  industrial perspective (2016)

\bibitem{wen2020convolutional}
Wen, J., et~al.: Convolutional neural networks for classification of
  alzheimer's disease: Overview and reproducible evaluation. MedIA
  \textbf{63},  101694 (2020)

\bibitem{wu2018registration}
Wu, Z., Li, G., Wang, L., Shi, F., Lin, W., Gilmore, J.H., Shen, D.:
  Registration-free infant cortical surface parcellation using deep
  convolutional neural networks. In: International Conference on Medical Image
  Computing and Computer-Assisted Intervention. pp. 672--680. Springer (2018)

\bibitem{xu2014weakly}
Xu, Y., Zhu, J.Y., Eric, I., Chang, C., Lai, M., Tu, Z.: Weakly supervised
  histopathology cancer image segmentation and classification. Medical image
  analysis  \textbf{18}(3),  591--604 (2014)

\bibitem{yang2021deep}
Yang, Z., et~al.: A deep learning framework identifies dimensional
  representations of alzheimer¡¯s disease from brain structure. Nature
  Communications  \textbf{12}(1),  1--15 (2021)

\bibitem{yuan2022explainability}
Yuan, H., et~al.: Explainability in graph neural networks: A taxonomic survey.
  IEEE TPAMI  (2022)

\bibitem{zhao2019spherical}
Zhao, F., et~al.: Spherical u-net on cortical surfaces: methods and
  applications. In: IPMI. pp. 855--866. Springer (2019)

\bibitem{zhao2022deep}
Zhao, F., et~al.: Deep learning in cortical surface-based neuroimage analysis:
  a systematic review. Intelligent Medicine  (2022)

\bibitem{zhou2016learning}
Zhou, B., et~al.: Learning deep features for discriminative localization. In:
  ICCV. pp. 2921--2929 (2016)

\bibitem{zhou2018brief}
Zhou, Z.H.: A brief introduction to weakly supervised learning. National
  science review  \textbf{5}(1),  44--53 (2018)

\end{thebibliography}

\end{document}